\documentclass{article}

\usepackage[preprint]{neurips_2021}

\usepackage{amsmath} 
\usepackage{graphicx} 

\usepackage[utf8]{inputenc} 
\usepackage[T1]{fontenc}    
\usepackage{hyperref}       
\usepackage{url}            
\usepackage{booktabs}       
\usepackage{amsfonts}       
\usepackage{nicefrac}       
\usepackage{microtype}      
\usepackage{xcolor}         

\usepackage{adjustbox}
\usepackage[algoruled,lined,ruled,resetcount,linesnumbered]{algorithm2e}
\usepackage{amssymb}
\usepackage{array}
\usepackage{caption}
\usepackage{colortbl}
\usepackage{enumitem}
\usepackage{etoolbox}
\usepackage{framed}
\usepackage{mathrsfs}
\usepackage{mathtools}
\usepackage{mdframed}
\usepackage{multicol}
\usepackage{multirow}
\usepackage{physics}
\usepackage{pifont}
\usepackage{placeins}
\usepackage{soul}
\usepackage{subcaption}
\usepackage[most]{tcolorbox}
\usepackage{titletoc}
\usepackage{wrapfig}
\usepackage{xltabular}
\usepackage{xspace}

\title{Reinforcement Learning for Hanabi}

\author{
 Nina Cohen\thanks{Equal Contribution}* \\
 Johns Hopkins University \\
 Baltimore, MD, USA \\
  \texttt{ncohen19@jh.edu} 
    \And
 Kordel France* \\
 Johns Hopkins University \\
 Baltimore, MD, USA \\
  \texttt{kfrance8@jh.edu} 
}

\begin{document}

\maketitle

\begin{abstract}
Hanabi has become a popular game for research when it comes to reinforcement learning (RL) as it is one of the few cooperative card games where you have incomplete knowledge of the entire environment, thus presenting a challenge for a RL agent. We explored different tabular and deep reinforcement learning algorithms to see which had the best performance both against an agent of the same type and also against other types of agents. We establish that certain agents played their highest scoring games against specific agents while others exhibited higher scores on average by adapting to the opposing agent's behavior. We attempted to quantify the conditions under which each algorithm provides the best advantage and identified the most interesting interactions between agents of different types. In the end, we found that temporal difference (TD) algorithms had better overall performance and balancing of play types compared to tabular agents. Specifically, tabular Expected SARSA and deep Q-Learning agents showed the best performance.
\end{abstract}

\section{Introduction}
Over the last few years, Hanabi has become an extremely popular game for reinforcement learning (RL) as it is not only a cooperative card game, but also a game where the players do not have complete knowledge of the environment. This combination leads to a more challenging game and a great problem for reinforcement learning. We explored the use of four different tabular and three deep RL (DRL) algorithms to see what type of agent or combination of agents performed the best. We leveraged an environment that had been previously created by Kantack, et al. (\cite{iai}) in order to be able to focus more on agent development. In the remainder of the paper, we will explore the card game Hanabi and discuss the different results from the four agents.

\section{Hanabi}
Hanabi is a cooperative card game where the goal is to stack similar colored cards in ascending order from one to five. What makes this game complicated is one cannot see their own cards, but they can see all the other player's cards and all players must work together to stack the cards. In order to learn what cards one has in their hand, one must rely on clues given by the other players indicating either a color or number currently in their hand. On a given turn, a player can make one of four moves: play a card, discard a card, give a color hint or give a number hint. The goal is that, for each of the five colors, the players will stack cards from one to five without losing all three of their life tokens. Tokens are lost by attempting to play a card that is not playable, meaning it is not the sequential card for its color's stack. The game is won by completing each color's stack, meaning successfully stacking cards up to five for each color. The game is lost if all life tokens are used or there are no more cards in the deck to be played. The final score is calculated by summing the number of cards in each of the center piles with a maximum score of 25. The deck contains for each color: three 1's, two 2's, two 3's, two 4's and one 5. We will discuss more in-depth the different turn options below.

\subsection{Playing a Card}
If a player selects to play a card on their turn, they will attempt to play the card on its corresponding color's pile. If it happens to be the next card that for that stack, it will be placed on top of the pile and that player will draw a new card without looking at the card's face. If the card cannot be played, then that player will lose a life token, the card will be placed at the top of the discard pile face up, and they will draw a new card without looking at its face. When one successfully plays a card, they are also granted a hint token, with a maximum number of hint tokens being 13. 

\subsection{Discarding a Card}
If a player selects to discard a card, they will place the discarded card on the discard pile face up and pick up a new card without looking at its face. Discarding a card allows the player to gain knowledge of what is left in the card pile, while also possibly remove non-playable cards in order to potentially obtain a playable card. When a player discards a card, they will also gain a hint token. 

\subsection{Hinting Color}
The first of the two types of hints is a color hint. Assuming the player has a hint token to use, a color hint entails telling one player one color they have in their hand. The player will then identify which cards in the opposing player's hand are that color so they can keep track of that information. This can be helpful for either singling out specific cards or giving the player more information about what they have in their hand. 

\subsection{Hinting Number}
The second type of hint is a number hint. Similar to a color hint, one provides another player with one number they have in their hand and identify which of their cards are that number. This allows them to have more information of which cards are playable, which to hold on to or which to discard.

\section{Solutions}
In this project, we used four tabular and four deep reinforcement learning algorithms and performed a number of experiments where agents would train against an agent of the same type and against an agent of a different type. For each of the tabular methods, we instantiated the deep learning version in order to be able to directly compare the performances of the two versions. Agents played 1,000 games and in order to be able to compare runs, we will be examining the following metrics: score per game, number of turns each agent took per game, number of times each agent played a card per game, number of times each agent discarded a card per game, and number of times each agent gave a hint per game. 

We performed three experiments:
\begin{enumerate}
    \item Evaluate the performance of two agents in learning to play the game, each of a different tabular TD method.
    \item Evaluate the performance of two agents in learning to play the game, each of a different deep learning TD method that is analogous to each tabular agent.
    \item Evaluate the effects of incrementally adding layers to the neural network of each of the two agents in attempt to find the optimal function approximation model.
\end{enumerate}

Throughout the experiment, we hypothesize that: 
\begin{enumerate}
\item Temporal difference reinforcement learning algorithms will perform better through self-play.
\item Games with at least one player as a tabular Expected SARSA agent will be the highest-scoring against all other tabular agent pairs in general due to Expected SARSA's ability to learn as both an on-policy and off-policy algorithm and promote longer games.
\item Games with both players as deep Q-learning agents will be both the shortest and the highest-scoring against all other deep learning agent types enabled partially by Q-learning's greediness parameter and the neural network's ability to perform effective function approximation.
\item Games where both players are deep learning agents will outperform a majority of games where both players are tabular agents when 2 or more hidden layers are leveraged within the neural network.
\end{enumerate}

\subsection{Reward}
As mentioned, the foundation of the Hanabi code was leveraged from the Instructive Artificial Intelligence (IAI) project from the Johns Hopkins Applied Physics Lab (\cite{iai}). We made the necessary alterations to enable the project to accommodate the TD agents described here and enable deep learning. For that project, rewards were calculated and whichever reward was the highest, the agent selected that action. We leveraged those rewards and used them as the reward that was given to agent in order to update the q table or neural network. This way, there were no static rewards for the 20 different moves and the state of the game was taken into account. The reward array that is returned is shape 20 x 12 where the 20 rows represent the 20 possible moves in the game and the 12 columns represent the different reasons the player could be making said move. Some of these are weighted higher than others as they lead to more favorable results or more pertinent information. The possible reward reason are as follows: 
\begin{enumerate}
\item Playing a card when you have more than 1 life tokens
\item Playing when you have 1 or less life tokens
\item Playing a singled out playable card
\item Providing a hint that singles out a playable card
\item Providing a hint that singles out a non-playable card
\item Discarding a singled out, playable card 
\item Providing hint on non-playable card
\item Providing hint on playable cards
\item If the card is 100\% playable, provide a playability bonus
\item Discarding a card in order to get a information token 
\item Safely discard a card that had hints on it and was safe to discard
\item Discard an unneeded card
\end{enumerate}

After every turn, the rewards were calculated and the reward for the selected action was returned to be used to update the table.

\subsection{Agent Types}
For our experiments, we leverage four temporal difference (TD) algorithms: Q-learning, SARSA, $n$-step SARSA, and Expected SARSA. Q-learning allows us to explore a greedy agent that will attempt to maximize its score early on in policy learning. SARSA emphasizes early exploration in constructing a policy and therefore may be advantageous in initially adapting to the other player's behavior. $n$-step methods allow an increased element of prediction by inferring \textit{n} steps into the future, another potential player advantage \footnote{Note that 1-step SARSA can be considered as the off-policy version of regular SARSA, inherently an on-policy method.}. Here, we evaluate step sizes of 1, 2, and 8. Expected SARSA affords us the freedom to fluctuate between an on-policy and off-policy method while providing a natural means of controlling reward loss. For our experiments, we allow $\epsilon$ (greediness) in Expected SARSA to increase as a function of the number of plays according to Sutton and Barto (\cite{Sutton2018}).

\subsection{Tabular Agents}
We hypothesized that Expected SARSA would be the best performing tabular RL algorithm in general. Since Expected SARSA seeks the \textit{expected} reward (or mean value), we suspected it would retain an improved ability to adapt to different agents behavior while bounding exploration. An agent seeking the average reward will be able to approximate the hidden true reward of an environment better than an agent that is always seeking the maximum reward, such as a Q-learner. The update rule for Expected SARSA is defined as the following:
\[
    Q^*(s,a) = Q(s, a) + \alpha \left [R(s, a, s') +  \frac{\gamma}{n}\sum_{i=1}^{n} Q(s'_i, a_i') - Q(s, a) \right ]
\]

Expected SARSA allows the algorithm to take advantage of both on-policy and off-policy training by moderating the learning of its policy as a function of correct reward assignment and the parameter $\epsilon$. 
In this fashion, early iterations of our Expected SARSA agent began with a policy that approximates on-policy SARSA,
\[
Q^*(s,a) = Q(s, a) + \alpha [R(s, a, s') + \gamma  Q(s', a') - Q(s, a)]\]
moderating the high loss, and then evolved into off-policy Q-learning
\[Q^*(s,a) = Q(s, a) + \alpha \left [R(s, a, s') + \gamma \  \max_{a'}\ Q(s', a') - Q(s, a)\right ]\]
as the policy became more refined \cite{vanSeijen2009}. This also afforded the algorithm to inherit the advantages of regular SARSA and Q-learning.

\subsection{Deep Learning Agents}
\begin{figure}[b]
  \centering
  \includegraphics[width=135mm]{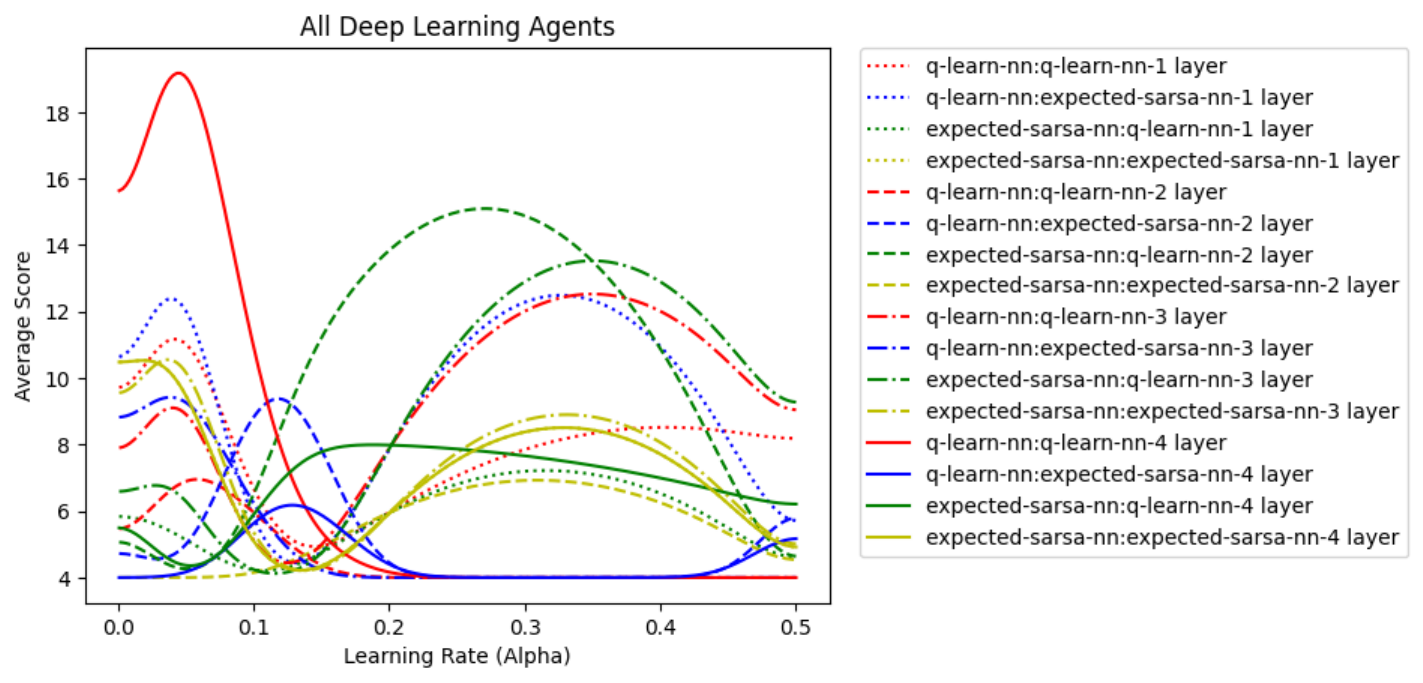}
  \caption{A summary of learning rate effects as the number of hidden layers increases.}
  \label{resultsAllNNLearningRatesA}
\end{figure}

Q-learning is an off-policy temporal difference algorithm that learns to approximate the optimal action-value function directly, independent of the policy being followed (\cite{Watkins1989}). It learns a policy that selects the next state-action pair that produces the maximum reward. It is one of the most commonly used TD algorithms in deep reinforcement learning due to its computational efficiency and good approximation of the total cumulative reward. We dismissed Q-learning as the top-performing tabular algorithm in favor of Expected SARSA due to the hypothesized ability of the latter to exhibit improved adaptability to other agent's behavior. However, we acknowledge that the neural network used in function approximation will enable the same behavior for a deep Q-learning agent, and that the inherent greediness of a Q-learner will maximize the reward achieved over all other deep learning agents.

To assess this hypothesis, we evaluated neural networks with 1, 2, 3, and 4 densely connected hidden layers for each agent. We use ReLU activation between hidden layers with a final Softmax output as the action classification. Mean squared error loss is used as the loss function and we leverage Adam as the optimizer, with $\beta_1$=0.900, $\beta_2$=0.999, $\epsilon_{adam}$ = 1e-07, and a momentum value $m$ = 0.990.

We declared different learning rates $\alpha \in [0.001, 0.5]$ and evaluated their effects on performance. We suspect a single hidden layer to not be large enough to capture the dimensionality of an effective Hanabi policy, so we expected Q-learning to be the superior algorithm in both maximum score achieved and highest average score against all individuals when 2 or more hidden layers are used. 

In order to establish a common learning rate and number of layers among all deep learning agents for the experiment, we performed an ablation study. Figure ~\ref{resultsAllNNLearningRatesA} shows the average score of 100 games for different agent matches. We only show the plot for Expected SARSA and Q-learning for clarity, but ultimately find that, over all agents, 4 hidden layers was and a learning rate of 0.01 was the most desireable architecture in general.

\section{Results}
We evaluated the results obtained from each of our agents in effectively learning to play the game of Hanabi and adapt to the opposing player's strategies. For each agent pair (e.g. SARSA vs. Q-learning), we played 1,000 games and averaged the results. We first summarized the performance of the tabular agents, then showed improvements made by the deep learning agents, and finally show results of the interaction between both agent classes playing against one another. In general, we found that the performance is maximized through function approximation and deep learning agents, albeit at the expense of increased computation.

\subsection{Results of Tabular Agents}
\begin{figure}[b]
  \centering
  \includegraphics[width=120mm]{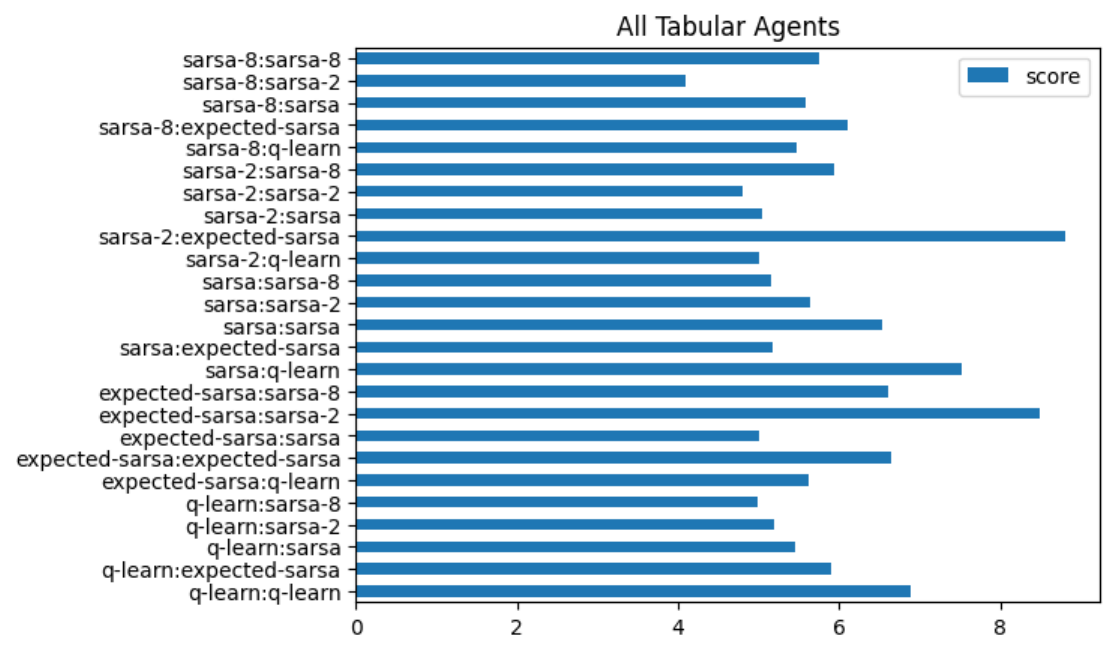}
  \caption{A summary of scores by each pair of tabular agents.}
  \label{resultsAllScoreA}
\end{figure}

Figure~\ref{resultsAllScoreA} illustrates the scores received by each agent pair throughout play. We acknowledge that games where at least one player is Expected SARSA indeed achieves highest average scores. Interestingly, when Expected SARSA played various $n$-step methods (denoted as \textit{expected-sarsa:sarsa-$n$}), we see higher scores when Expected SARSA played first, but not when it played second. 

\begin{figure}[t]
  \centering
  \includegraphics[width=120mm]{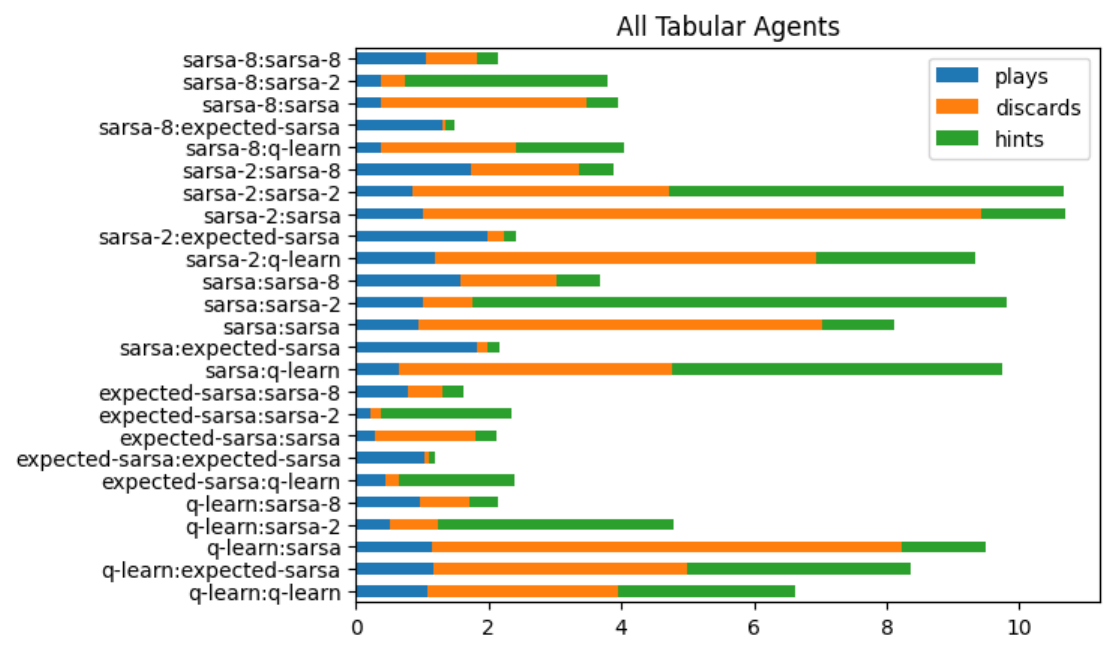}
  \caption{A summary of turns by each pair of tabular agents.}
  \label{resultsAllTurnsA}
\vspace{-5mm}
\end{figure}

The distribution of turns is shown in Figure~\ref{resultsAllTurnsA}. Games where Expected SARSA played first showed much lower turn counts than those games where it played second. One can observe that the biggest contributor to the number of turns when, for example, the 2-step SARSA agent played first is the number of discards, but when the Q-learning agent played first, the number of hints is generally the biggest contributor. 

When it comes to agents move types versus their final score, commonly you expect to see a large percentage of the moves be plays as playing the card is only way to gain points. In a perfect game, there would be at least 25 plays in order to get all of the cards stacked in the center. If you see a large, equal number of hints and plays, that could possibly indicate one agent consistently hinting to the other of their playable cards and the second agent dominating playing cards. The lower the number of turns where agents played cards, the lower the score will be overall as that is the only way to obtain points. If there is significantly more plays than hints, as seen with \textit{sarsa-2:sarsa}, that indicates that one or both agents are more risky and willing to play cards they know very little about.

In summary of the tabular agent results, we found evidence to support our hypothesis that Expected SARSA agents contributed to the highest scoring games. However, we also found supporting evidence against the hypothesis since the Expected SARSA agents shared the highest average score with other agent pairs, many including various $n$-step SARSA methods. 

\subsection{Results of Deep Learning Agents}

\begin{figure}[b]
  \centering
  \includegraphics[width=120mm]{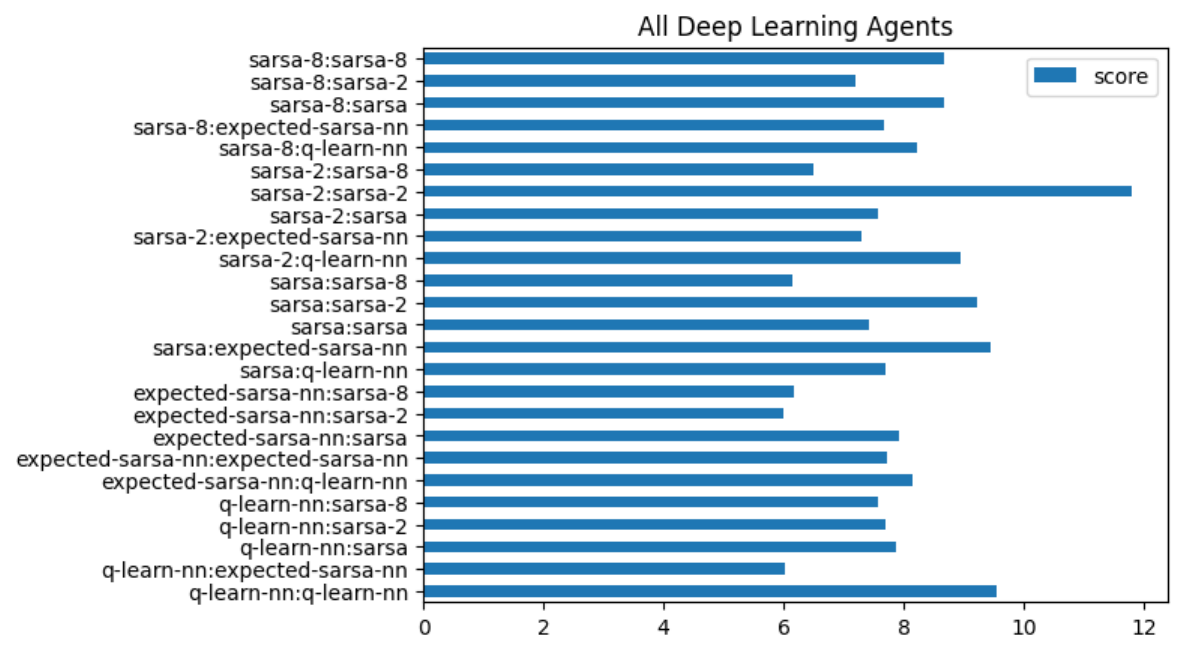}
  \caption{A summary of scores by each pair of deep learning agents.}
  \label{resultsAllNNScoreA}
\end{figure}

\begin{figure}
  \centering
  \includegraphics[width=120mm]{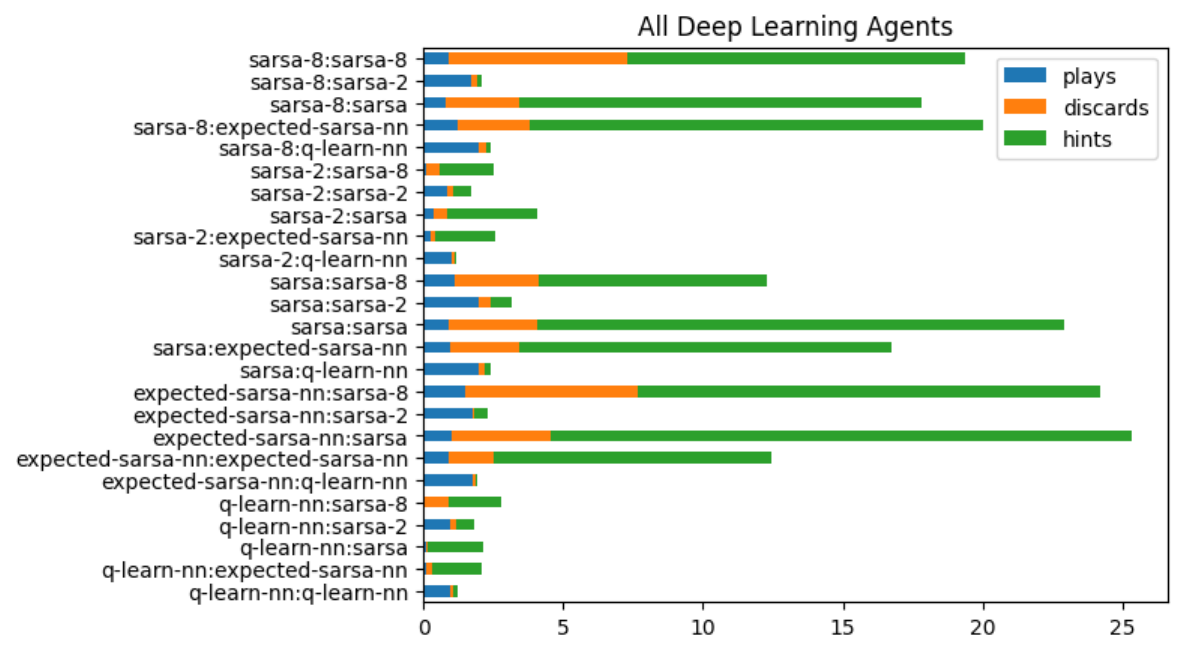}
  \caption{A summary of turns by each pair of deep learning agents.}
  \label{resultsAllNNTurnsA}
\end{figure}

Figures ~\ref{resultsAllNNScoreA} and ~\ref{resultsAllNNTurnsA} show synonymous plots with the previous section, but focus on the performance of the deep learning agents. Scores, on average, are higher for the deep learning agents than for the tabular agents. We observed that games with both 2-step SARSA agents return the highest scores and games where both players were Q-learning agents return the second highest scores. 

Additionally, the number of discards, hints, and plays were markedly different in proportion to the each agent's tabular counterpart. There were much fewer discards with deep learning than with tabular methods. Games where at least one of the players was either a Q-learning agent or 2-step SARSA agent were also lower scoring, which may indicate that these agent types significantly excelled at the game in comparison to their opponent and therefore promoted a shorter game.

\section{Discussion}
In consideration of our results, we found substantial evidence to support our first hypothesis that deep TD reinforcement learning algorithms can indeed learn to play the game of Hanabi effectively. We also found clear support for hypotheses 2 and 3, but that support was not unanimous. Expected SARSA proved to be the most effective tabular method, while Q-learning and 2-step SARSA emerged as the most effective deep learning methods when 4 hidden layers were used. This may give credence to $n$-step SARSA's ability to "look ahead" and infer the state-action values. One reason why the superior performance of Expected SARSA did not carry over into the deep learning could be due to the fact that the neural network was performing function approximation that replaced Expected SARSA's advantage in seeking the expected reward. Expected SARSA acts as a function approximator for tabular methods, but our results give evidence that this could be replaced by the more accurate neural network in deep learning. 

The benefits that a neural network can provide in learning to play Hanabi through function approximation are apparent and this supports our final hypothesis. For 80\% of agents, we saw score improvements with the deep learning variant with a p-value of 0.0038 using the Wilcoxon-Signed Rank Test. With this, we established more evidence to support hypothesis 4, suggesting that the use of function approximation by an agent may be an indication of better learning in general. We note, however, that the results between tabular and neural network agents were fairly comparable. This begs the question of whether the improved performance gained by deep learning is worth the large increase in computational time and resources. On average, we found that training times for each of the deep learning games took 8-12x longer to train on a single core hyper-threaded Intel Xeon processor with 2.3 Ghz frequency and 3-4x longer when also using a single Nvidia Tesla K80 GPU. While wall times of an algorithm should never be used as a meticulous measure of the its computational complexity, we use it here only as a proxy to assess the trade-off needed  to utilize a deep learning enabled agent.

In future work, we hope to examine how tabular and deep agents play against one another, while also looking at the effects of inverse reinforcement learning and imitation learning (\cite{ciosek2022imitation}). Similar to the manner in which the temporal difference methods above performed better when coupled with certain agents over others, we expect an inverse reinforcement learning agent will need to effectively generalize to different strategies since different humans play with different idiosyncrasies.

\section{Conclusion}
In conclusion, we were able to complete three Hanabi experiments where we tested the performance of four tabular and four deep reinforcement learning algorithms. In the end, when Expected SARSA was at least one player in the game, it had the best performance for the tabular agents. Q-learning and 2-step SARSA proved most effective for the deep reinforcement algorithms. DRL agents also had significantly more hints than tabular, while also achieving higher scores. Additional work can be done to see how deep and tabular agents play against one another, however in this project we were able to prove that tabular and deep reinforcement algorithms have fairly comparable results in the card game Hanabi.

\section*{Acknowledgements}
Nina Cohen and Kordel France performed this work as partial fulfillment for the Master of Science program in Artificial Intelligence under instruction of Dr. Mark Happel at Johns Hopkins University.

\bibliography{bibs.bib}

\begin{thebibliography}{5}
\providecommand{\natexlab}[1]{#1}
\providecommand{\url}[1]{\texttt{#1}}
\expandafter\ifx\csname urlstyle\endcsname\relax
  \providecommand{\doi}[1]{doi: #1}\else
  \providecommand{\doi}{doi: \begingroup \urlstyle{rm}\Url}\fi

\bibitem[Ciosek(2022)]{ciosek2022imitation}
Kamil Ciosek.
\newblock Imitation learning by reinforcement learning.
\newblock In \emph{International Conference on Learning Representations}, 2022.

\bibitem[Kantack et~al.(2022)Kantack, Cohen, Bos, Lowman, Everett, and
  Endres]{iai}
Nicholas Kantack, Nina Cohen, Nathan Bos, Corey Lowman, James Everett, and Tim
  Endres.
\newblock Instructive artificial intelligence (ai) for human training,
  assistance, and explainability.
\newblock In \emph{Artificial Intelligence and Machine Learning for
  Multi-Domain Operations Applications IV}, volume 12113, pages 45--54. SPIE,
  2022.

\bibitem[Sutton and Barto(2018)]{Sutton2018}
Richard~S. Sutton and Andrew~G. Barto.
\newblock \emph{Reinforcement Learning: An Introduction}.
\newblock The MIT Press, second edition, 2018.

\bibitem[van Seijen et~al.(2009)van Seijen, van Hasselt, Whiteson, and
  Wiering]{vanSeijen2009}
Harm van Seijen, Hado van Hasselt, Shimon Whiteson, and Marco Wiering.
\newblock A theoretical and empirical analysis of expected sarsa.
\newblock In \emph{2009 IEEE Symposium on Adaptive Dynamic Programming and
  Reinforcement Learning}, pages 177--184, 2009.

\bibitem[Watkins(1989)]{Watkins1989}
Christopher Watkins.
\newblock \emph{Learning from Delayed Rewards}.
\newblock PhD thesis, Department of Computer Science, King's College, Cambridge
  University, 1989.

\end{thebibliography}
\bibliographystyle{plainnat}

\end{document}